\documentclass{article}

\usepackage[final]{neurips_2021}

\usepackage[utf8]{inputenc} 
\usepackage[T1]{fontenc}    
\usepackage{hyperref}       
\usepackage{url}            
\usepackage{booktabs}       
\usepackage{amsthm, amssymb, amsmath, latexsym, amsfonts, mathcomp}       
\usepackage{nicefrac}       
\usepackage{microtype}      
\usepackage{xcolor}         
\usepackage{graphicx}

\def\S2{{\mathrm{S}^2}}

\title{Towards Representation Learning for Atmospheric Dynamics \\[5pt]
\large{Effective Distance Measures to Address Climate Change}}

\author{%
  Sebastian Hoffmann \\
  Dept. of Computer Science\\
  Universit{\"a}t Magdeburg\\
  \texttt{sebastian1.hoffmann@ovgu.de} \\
 \And
 Christian Lessig \\
  Dept. of Computer Science\\
  Universit{\"a}t Magdeburg\\
  \texttt{christian.lessig@ovgu.de}
}

\begin{document}

\maketitle

\begin{abstract}
  The prediction of future climate scenarios under anthropogenic forcing is critical to understand climate change and to assess the impact of potentially counter-acting technologies. 
  Machine learning and hybrid techniques for this prediction rely on informative metrics that are sensitive to pertinent but often subtle influences.
  For atmospheric dynamics, a critical part of the climate system, no well established metric exists and visual inspection is currently still often used in practice. 
  However, this ``eyeball metric'' cannot be used for machine learning where an algorithmic description is required.
  Motivated by the success of intermediate neural network activations as basis for learned metrics, e.g. in computer vision, we present a novel, self-supervised representation learning approach specifically designed for atmospheric dynamics.
  Our approach, called AtmoDist, trains a neural network on a simple, auxiliary task: predicting the temporal distance between elements of a randomly shuffled sequence of atmospheric fields (e.g. the components of the wind field from reanalysis or simulation). 
  The task forces the network to learn important intrinsic aspects of the data as activations in its layers and from these hence a discriminative metric can be obtained.
  We demonstrate this by using AtmoDist to define a metric for GAN-based super resolution of vorticity and divergence. 
  Our upscaled data matches both visually and in terms of its statistics a high resolution reference closely and it significantly outperform the state-of-the-art based on mean squared error.
  Since AtmoDist is unsupervised, only requires a temporal sequence of fields, and uses a simple auxiliary task, it has the potential to be of utility in a wide range of applications.
\end{abstract}

%

\section{Introduction}

A discriminative distance measure for atmospheric dynamics is critical for a wide range of applications addressing climate change.
Such a measure could, for instance, enable climate scientists to automatically classify patterns, such as polar jet behavior or blocking events, and how these change under anthropogenic forcing.
It is also an amplifying technology for many machine learning techniques in the context of climate change since it allows for more effective loss functions or evaluation metrics.
A third potential application, and the one motivating the present work, is the use in a hybrid climate simulations where a classical discretization is combined with a neural network that corrects for model biases and represents unresolved scales.

Motivated by the deficiencies of existing distance measures~\citep{Stengel2020}, we introduce in the present work AtmoDist, a representation learning approach tailored towards atmospheric dynamics that uses intermediate neural network activations to obtain an effective metric for such data.
AtmoDist only requires time series of one or more atmospheric fields, e.g. from reanalysis or a simulation, as input and learns from it, through an appropriate choice of fields, a domain specific distance measure.
We demonstrate the utility of AtmoDist by using it for GAN-based super resolution of vorticity and divergence, i.e. the potentials for the atmospheric wind velocity field.
Results obtained with AtmoDist match visually and in terms of key statistics the ground truth significantly more closely than mean squared error, as used, e.g., in the state-of-the-art~\citep{Stengel2020}.



\section{AtmoDist}
\label{sec:AtmoDist}

\begin{figure}[t]
  \centering
  \includegraphics[width=0.85\textwidth]{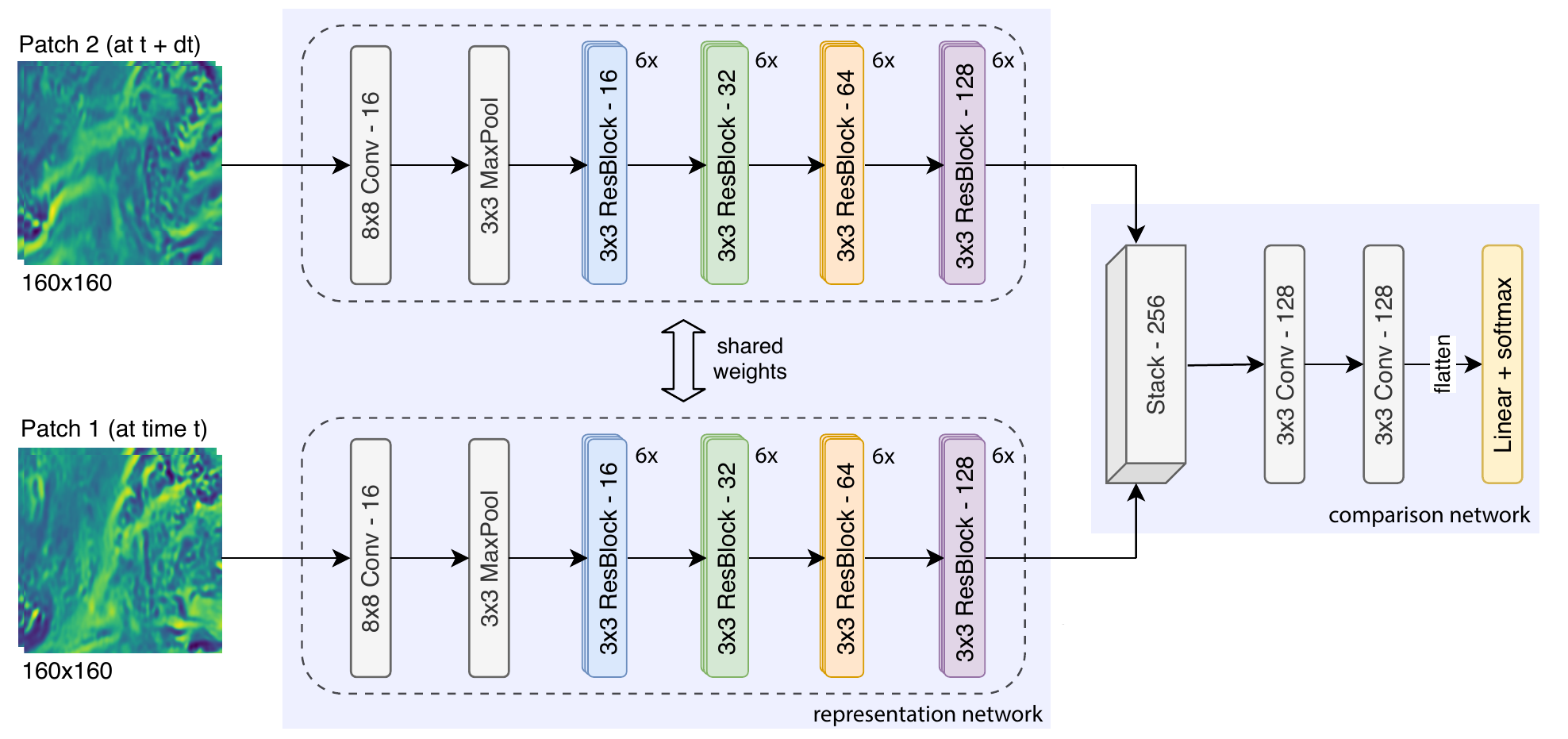}
  \caption{Network architecture used for AtmoDist. The representation network has a siamese structure and generates the intermediate representations for the input signal that distill important features. The comparison network is necessary only during training.}
  \label{fig:network}
\end{figure}

Mathematical distance measures, such as mean square error (MSE), are often not well suited as neural network loss function since they are oblivious to the relative importance of specific characteristics of the data for an application.
For atmospheric data, this has recently been pointed out by~\cite{Stengel2020}.
More advanced statistical measures, such as structural similarity index measure (SSIM), can alleviate some of the problems but they are still largely ignorant to an application.
As an alternative, intermediate neural network representations were used previously, e.g., in computer vision, to obtain more suitable metrics~\citep{heusel2017gans,salimans2016improved}, e.g. for SOTA super-resolution~\citep{SRGAN}.

Inspired by the success in computer vision, we propose a representation learning approach specifically designed for atmospheric data. 
To avoid the need for labelled training data, which is currently limited and whose generation is highly challenging, we use an unsupervised learning approach and exploit that, at least for short times, the temporal distance is a useful proxy for the intrinsic distance of the dynamics (indeed, in the case of an ideal atmospheric flow, the intrinsic distance between flow states is proportional to their time difference~\citep{Arnold1966,Ebin1970}). 
We hence use the identification of the (categorial) time difference between two temporally nearby inputs as auxiliary pretext task.
This forces the representation network to learn informative internal representations of the input, given through the activations at intermediate layers~\citep{Krizhevsky2012}.
The difference between internal activations, typically at one of the last layers, for two different inputs then provides a domain-specific distance measure (more precisely, a norm, such as MSE, of the difference).

\paragraph{Network architecture}

The network architecture of AtmoDist is shown in Fig.~\ref{fig:network}. It has the structure of a siamese network~\citep{Chicco2021} with two inputs being fed through the same residual representation network~\citep{he2016deep}.
The resulting weights are then stacked and analyzed in the comparison network that produces the categorial prediction of the temporal difference between the input signals.

The representation network is loosely inspired by~\cite{he2016deep} for classification on CIFAR-10~\citep{krizhevsky2009learning}.
It differs from configurations employed for classification on ImageNet mainly through the reduced number of parameters ($2.27 \times 10^6$ parameters for our network compared to $10.99 \times 10^6$ for ResNet-18, the smallest one proposed by~\cite{he2016deep}). We found that larger networks resulted in severe overfitting on our dataset. 
The comparison network effectively acts as a learnable metric taking in the two intermediate representations and mapping these to their temporal distance. We found that its use, as opposed to classifying directly on the stacked representations, results in a significantly increased classification accuracy.

\paragraph{Training for vorticity and divergence}

To evaluate the proposed representation learning network, we train it on (relative) vorticity and divergence, which together provide the potentials for the atmospheric wind velocity vector field.
The training data is obtained from ERA5 reanalysis~\citep{era5}, which we assume to be ground truth, measured atmospheric state.
Training is performed on the time period from 1979 to 1998 (20 years) while the period from 2000 to 2005 is used for evaluation (6 years)
This results in $58,\!440$ spatial fields for the training set and $17,\!536$ ones for the evaluation set.
The fields are sampled on a regular latitude-longitude grid with resolution $1280 \times 2560$ and at approximately $883 \, \mathrm{hPa}$ height.
Training is performed on randomly sampled patches of $160 \times 160$ pixels (about $2500 \, \mathrm{km} \times 2500 \, \mathrm{km}$) with a maximum time difference of $69 \, \mathrm{h}$.
The data follows a zero-centered Laplace distribution, i.e. it irregularly takes on very high values (relative to the standard deviation). 
To improve the stability and quality of the training we therefore $\log$-transform vorticity and divergence in a preprocessing step (see the appendix for details).

\begin{figure}[t]
  \centering
  \includegraphics[width=\textwidth]{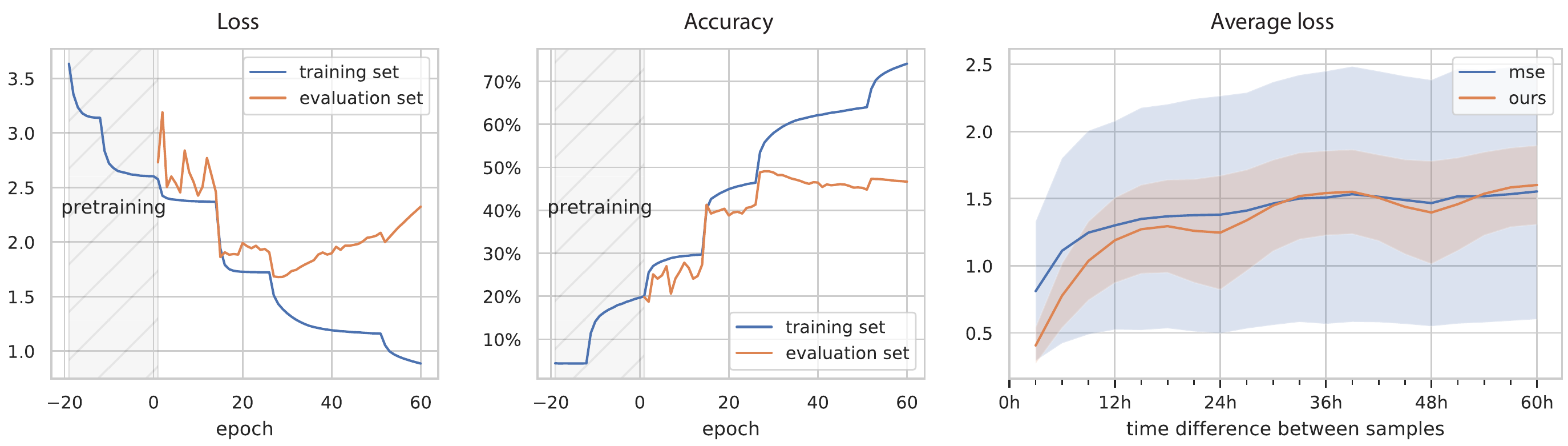}
  \caption{\emph{Left and middle:} Loss and accuracy of AtmoDist trained on vorticity and divergence. \emph{Right:} Average loss and standard deviation (shaded) on the training set with a distance measure derived from our trained representation network ($L_2$ norm of activation differences in last layer) and for MSE loss.}
  \label{fig:rptrain}
\end{figure}

The training loss and accuracy as a function of the training epoch is shown in Fig.~\ref{fig:rptrain}, left and middle. 
The results demonstrate good convergence of the training with the performance on the evaluation set improving up to epoch $27$; afterwards overfitting sets in.
Fig.~\ref{fig:rptrain}, right, shows the average loss for the training set where the activations of the last layer of the trained representation network (violet block in Fig.~\ref{fig:network}) is used as distance measure, as described above.
The results demonstrate  the effectiveness of our ansatz since a more linear increase in the distance and lower variance is achieved. 

%

\section{AtmoDist for Super Resolution}
\label{sec:super_resolution}

Super resolution, or downscaling, is a classical problem in climate science. 
It arises from the lack of scale closure, i.e. while there is a coarsest scale, given by large scale patterns stretching over thousands of kilometers, there is no smallest one for climate dynamics and with each new scale, down to meters and centimeters, additional phenomena arise.
Traditionally, this leads to larger and larger computers being used, with exascale computing being required to reach the $10 \, \mathrm{km}$ grid resolution that is the state-of-the-art~\citep{Schulthess2019}.
Recently, GANs have been proposed as an alternative to downscale both coarse scale simulations or observational data, e.g.~\citep{Stengel2020,Kashinath2021}.
We will demonstrate the practical utility of AtmoDist with the state-of-the-art in GAN-based downscaling~\citep{Stengel2020} and comparing the MSE loss used in this work with a AtmoDist based representation loss. 

\begin{figure}[t]
  \centering
  \includegraphics[width=\textwidth]{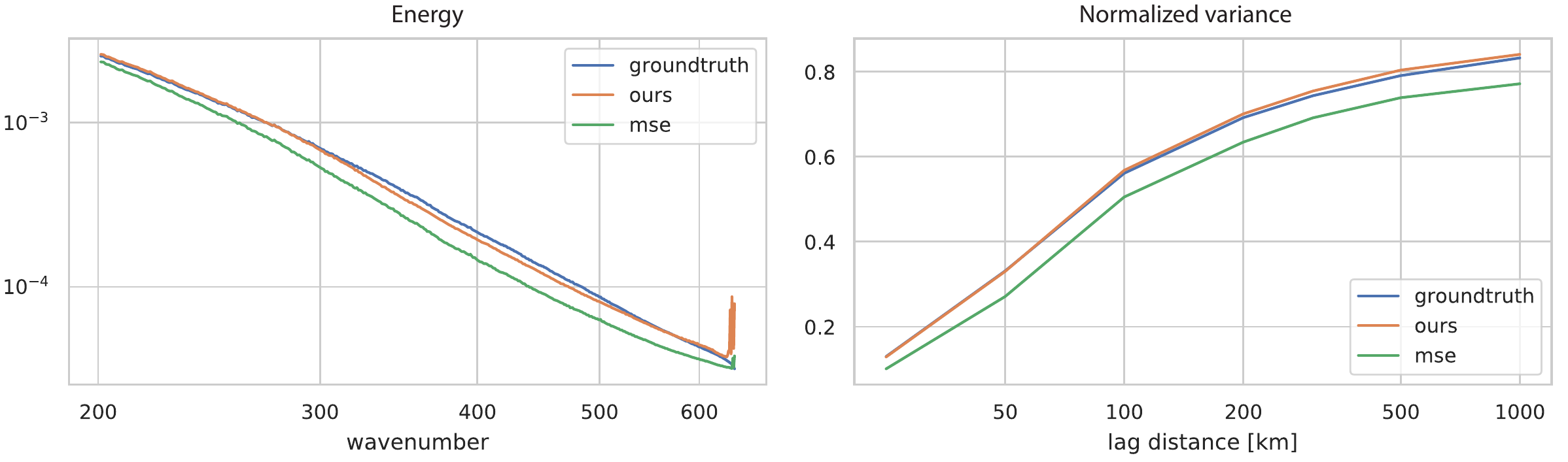}
  \caption{Energy spectrum (left) and vorticity semivariogram (right) for our representation learning approach and MSE loss. The semivariogram has been normalized to the evaluation set variance.}
  \label{fig:superres_stats}
\end{figure}

\begin{figure}[b]
  \centering
  \includegraphics[width=\textwidth]{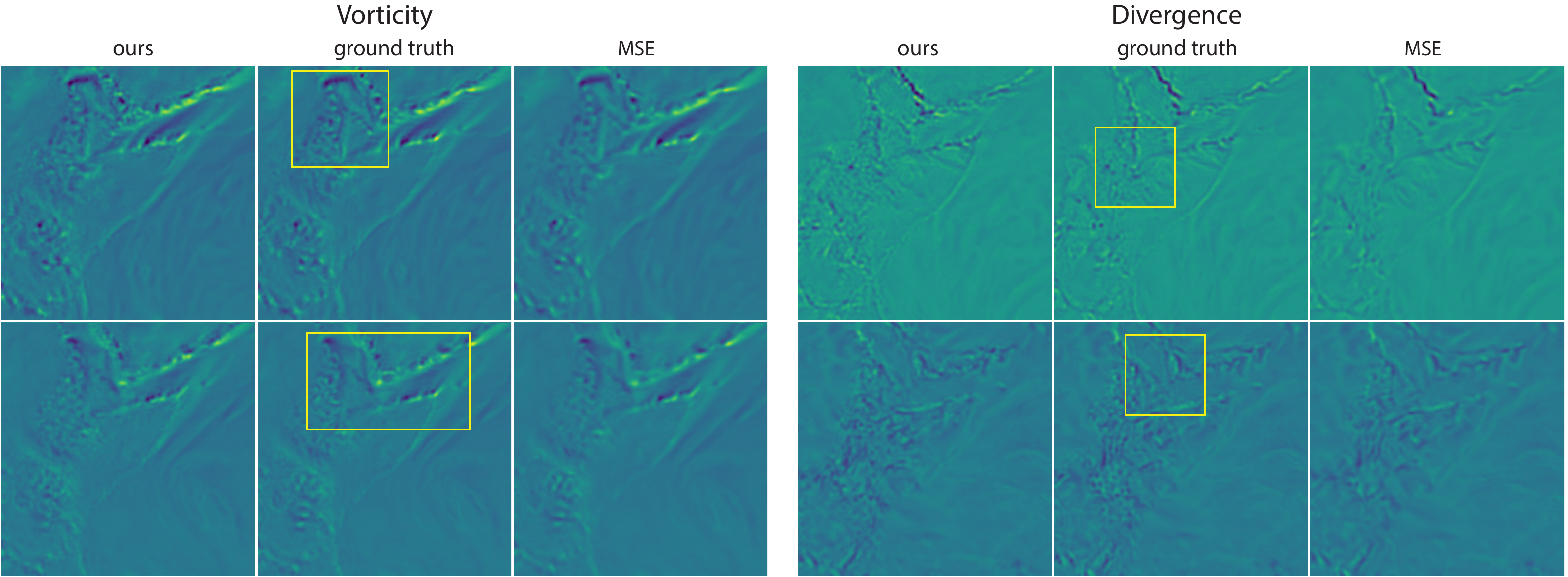}
  \caption{Uncurated super resolution examples for vorticity (left) and divergence (right). The yellow boxes highlight regions where the difference between AtmoDist and MSE is particularly apparent.}
  \label{fig:superres_spatial}
\end{figure}

Results for $4$x super resolution are presented in Fig.~\ref{fig:superres_stats} and Fig.~\ref{fig:superres_spatial}.
The energy spectrum in Fig.~\ref{fig:superres_stats}, left, shows a clear improvement compared to MSE and almost perfectly fits the ground truth. 
Some artifacts are apparent for the highest frequencies but these can easily be suppressed by an appropriate low pass filter.
In Fig.~\ref{fig:superres_stats}, right, we show the semivariogram~\citep{matheron1963principles} for vorticity, which measures geospatial autocorrelation. 
It verifies that our approach captures geostatistical properties significantly better than MSE; similar results are obtained for divergence.
Fig.~\ref{fig:superres_spatial} shows examples for the generated super resolution. 
Clearly visible are the overly smooth results obtained with MSE and that our representation network based loss significantly improves this.

\section{Discussion}
\label{sec:discussion}

We introduced AtmoDist, a representation learning approach tailored towards atmospheric dynamics that only requires a time series to obtain a domain specific distance measure for atmospheric data.
We demonstrated the utility of our approach by using AtmoDist for super resolution of vorticity and divergence.
Compared to MSE used in the state-of-the-art~\citep{Stengel2020}, we obtained significantly improved results both for the visual error and in terms of the statistics of the generated fields.
More details on AtmoDist will be presented in a forthcoming publication; the code is available on github.\footnote{\url{https://github.com/sehoffmann/AtmoDist}}

In future work, we would like to evaluate our ansatz more thoroughly, e.g. by considering physical fields other than vorticity and divergence, integrate vertical dependence, perform an ablation study to better understand design choice, perform hyper-parameter tuning, and thoroughly evaluate which block of the representation network is best suited to obtain a distance measure in different applications. 
An important direction in future work is also the application of AtmoDist to address climate change.
For example, the automatic classification of polar jet behavior (e.g.~\cite{Woollings2010}) or of blocking events (e.g.~\cite{Davini2012}) could be used to better understand the effect of global heating. 
Our long term objective is to develop a hybrid climate simulation that combines a traditional simulation based on partial differential equations with a neural network based correction.
AtmoDist developed in the present paper is a first but crucial step towards this objective.



\bibliography{references}


\appendix

\section{Appendix}
\label{sec:appendix}

\paragraph{Network architecture}

The representation network begins with a strided $8 \times 8$ convolution followed by a strided $3 \times 3$ max-pooling operation. Each residual block depicted in Fig.~\ref{fig:network} consists of two convolutions and a skip-connection (see \citep{he2016deep} for details). After doubling the number of channels in each layer, the spatial resolution is halved using strided convolutions.

Furthermore, each convolution (in a residual block or standalone) is followed by a batch~normalization~layer~\citep{ioffe2015batch} before being passed to a ReLU activation function~\citep{he2015delving}. For skip connections that operate across two different channel dimensions, learnable linear projections are employed. All non-bias weights are initialized according to the method described in~\cite{he2015delving}.

For super-resolution, we build upon the SRGAN~\citep{SRGAN} implementation made openly available by \cite{Stengel2020}. Our only modifications are an incorporation of the improved initialization scheme for the upscaling sub-pixel~convolutions~\citep{aitken2017checkerboard} as well as replacing transposed convolutions with regular ones in the generator as in the original architecture~\citep{SRGAN}. In particular, we keep the omission of batch~normalization~layers in the generator introduced by \cite{Stengel2020}.

\paragraph{ERA5 training data}

Our training data is computed onto a regular latitude-longitude grid with resolution $1280 \times 2560$ directly from the native spherical harmonics coefficients of ERA~5 at model level 120. 
Due to their Laplace distribution, the vorticity and divergence input data is $\log$-transform by $y=f(x)$ with the following mapping
\begin{align*}
    y = \frac{w - \mu_2}{\sigma_2} \qquad
    w = \mathrm{sign}(z) \log(1 + \alpha \left\vert z \right\vert) \quad
    z = \frac{x - \mu_1}{\sigma_1} \qquad
\end{align*}
which is applied element-wise and channel-wise.
Here $\mu_1$ and $\sigma_1$ denote the mean and standard deviation of the corresponding input channel, respectively, while $\mu_2$ and $\sigma_2$ denote the mean and standard deviation of the $\log$-transformed field $w$. All moments are calculated across the training dataset.
The parameter $\alpha$ controls the strength by which the dynamic range at the tails of the distribution is compressed.\footnote{Notice that the $\log$ function behaves approximately linear around 1, thus leaving sufficiently small values almost unaffected.}
We found that $\alpha = 0.2$ is sufficient to stabilize training while minimizing the compression applied to the original data. 

For the representation learning network, $21$ vorticity-divergence patch pairs of size $160 \times 160$ each are sampled randomly from the $1280 \times 2560$ fields at each time step in the input data. Due to the distortion near the poles, the latitude of the patch center is restricted to $60 \, ^{\circ}N$ to $60 \, ^{\circ}S$. This results in patches starting between $82.5 \, ^{\circ}N$ and $60 \, ^{\circ}N$ to still be included yet with linearly decreasingly probability; analogously in the Southern hemisphere. To obtain input for the super-resolution, we employ the same sampling scheme but sample $180$ patches of size $96 \times 96$ per time step. The low-resolution versions for these patches is then found by average pooling.

\paragraph{Training details}

The representation network is trained using standard stochastic gradient descent with momentum $\beta = 0.9$ and an initial learning rate of $\eta = 10^{-1}$. If training encounters a plateau, the learning rate is reduced by an order of magnitude to a minimum of $\eta_\text{min} = 10^{-5}$. Additionally, gradient clipping is employed, ensuring that the $l_2$-norm of the gradient does not exceed $5.0$. To counteract overfitting, weight decay of $10^{-4}$ is used. The network is trained with regular $\log$-loss, i.e. $L = -\frac{1}{N}\sum_{i=1}^N\mathbf{y}_i \log(\mathbf{\hat{y}}_i)$, where $\mathbf{y}_i$ is a vector with one-hot encoded class-labels and $\mathbf{\hat{y}}_i$ are the predictions of the network.

In preliminary experiments with lower resolutions we observed that training is initially very slow for the first four to five epochs before suddenly progressing rapidly.
Furthermore, in initial test with the full resolution of $1280 \, \times \, 2560$ the training did not converge at all.
We hypothesize that this behavior stems from the difficulty of the task combined with an initial lack of discerning features. We thus employ a pre-training scheme inspired by curriculum~learning~\citep{curriculum} where initially simpler examples are presented to the machine learning algorithms before gradually increasing the difficulty.

In particular, we performed the training of the network first on a smaller subset of the data. 
While eventually leading to severe overfitting if continued for too long, some of the lower-level features learned by the network generalize and prove useful for the complete dataset. Concretely, the network is pre-trained for 20 epochs on a fixed subset of 2000 batches before the learning rate is reset to $\eta=10^{-1}$ and training on the whole dataset begins. This pre-training scheme was sufficient to ensure convergence of the training as shown in Fig.~\ref{fig:rptrain}.

For super-resolution, we use the same training parameters as \cite{Stengel2020}. In particular, the learning rate is set to $\eta = 10^{-4}$ and the adversarial loss component is scaled by $\alpha_\text{adv} = 10^{-3}$. The GANs (one for MSE loss and one for our representation loss) are then trained for $6$ epochs using Adam~\citep{kingma2017adam}.

Training the representation network took approximately $1.5$ days on a NVIDIA RTX 2080 Ti, while training a single GAN on a NVIDIA Quadro RTX 6000 took approximately $4$ days.

\paragraph{Scaling of the loss function}

To ensure that MSE-based and AtmoDist-based super-resolution exhibit the same training dynamics, we normalize our loss-function. 
This is particularly important with respect to the $\alpha_\text{adv}$ parameter which controls the trade-off between content-loss and adversarial-loss.

We hypothesize that due to the chaotic dynamics of the atmosphere, any loss function should on average converge to a specific level after a certain time period (ignoring daily and annual oscillations, compare Fig.~\ref{fig:rptrain}, right). Thus, we normalize our content-loss by ensuring that these equilibrium levels are roughly the same in terms of least squares by solving the following optimization problem for the scaling factor $\alpha_\text{cnt}$
\begin{equation}
\underset{\alpha_\text{cnt} \in \mathbb{R}}{\text{minimize}} \ \sum_{t={\lfloor}N/2{\rfloor}}^N (\alpha_\text{cnt} \mathbf{c}_t - \mathbf{m}_t)^2
\end{equation}
where $\mathbf{c}_t$ denote the average content-loss of samples that are $t$ (categorical) timesteps apart, and $\mathbf{m}_t$ denote the average MSE of samples that are $t$ (categorical) timesteps apart respectively (compare Fig.~\ref{fig:rptrain}, right). It is easy to verify that the above optimization problem has the unique solution
\begin{equation}
  \alpha_\text{cnt} = \frac{\sum_{t={\lfloor}N/2{\rfloor}}^N \mathbf{c}_t \mathbf{m}_t} {\sum_{t={\lfloor}N/2{\rfloor}}^N \mathbf{c}_t^2} .
\end{equation}

\end{document}